[Type here]

PGP in Data Science

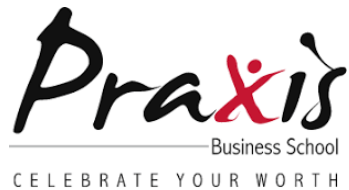

# A Comparative Analysis of Portfolio Optimization Using Mean-Variance, Hierarchical Risk Parity, and Reinforcement Learning Approaches on the Indian Stock Market

Capstone project report submitted in partial fulfilment of the requirements for the Post Graduate Program in Data Science at Praxis Business School

By

**Aditya Jaiswal (A22003)**

**Anshuman Pathak (A22008)**

**Atish Kumar Majee (A22012)**

**Kushagra Kumar (A22021)**

**Manas Kumar Sarkar (A22022)**

**Soubhik Maji (A22032)**

Under the supervision of

**Prof. Jaydip Sen**

**Professor, Praxis Business School**

[Type here]

# CONTENTS

# 1. Preliminaries
## 1.1. Nifty50

Nifty50 is a stock market index in India that represents the performance of the top 50 companies listed on the National Stock Exchange (NSE) of India. It is considered as a benchmark index for the Indian stock market and is closely watched by investors and analysts. The Nifty50 index includes companies from various sectors such as finance, technology, energy, healthcare, consumer goods, and more. Some of the well-known companies included in the index are Reliance Industries, HDFC Bank, Infosys, Tata Consultancy Services, and Hindustan Unilever.

The volatility of the stock market refers to the degree of fluctuation in stock prices over a certain period of time. Higher volatility means that stock prices are fluctuating rapidly and unpredictably, while lower volatility means that stock prices are relatively stable.

The Nifty50 index is also subject to volatility, as it is influenced by various factors such as global economic conditions, political events, and company-specific news. Factors such as interest rates, inflation, and government policies can also impact the volatility of the Nifty50 index.

Investors should be aware of the volatility of the Nifty50 index and the stock market in general, as it can affect their investment decisions and portfolio performance. High volatility may provide opportunities for high returns but also involves higher risks, while low volatility may provide stability but may offer lower returns. Therefore, it is important to consider individual investment goals, risk tolerance, and time horizon before making investment decisions in the stock market.

## 1.2 Risk and Return

The risk and return of a stock are important concepts in finance that describe the potential gain or loss associated with investing in a particular stock.

Risk (denoted as "σ ") is a measure of the uncertainty or variability of returns. There are several ways to calculate risk, and one common method is to use standard deviation, which measures the dispersion of returns around the average. The formula for calculating risk using standard deviation is:

$$\sigma^2 = \Sigma \frac{(x - \mu)^2}{N}$$

Where:

σ: Risk

$X_i$: Individual return for each period

μ: Average return

N: Number of periods

Return (denoted as "$r_j$") is the profit or loss generated from an investment, expressed as a percentage of the original investment. The formula for calculating return is:

$$r_j = \frac{P_{t+1} - P_t}{P_t} \equiv \frac{P_{t+1}}{P_t} - 1$$

Where Pt & Pt+1 is the Price of the stock at time t and at time t+1 respectively.

Investors generally expect a higher return for taking on higher levels of risk. However, risk and return are inherently linked, as investments with higher potential returns often come with higher levels of risk. It is important for investors to carefully assess both the risk and return characteristics of a stock or any investment before making investment decisions.

## 1.3 Portfolio & its optimization

A portfolio is a collection of financial assets, such as stocks, bonds, and other investments, held by an individual or an entity.

We need portfolio optimization because investing in a single asset or stock can be risky. Investing in a portfolio of assets, on the other hand, can help to reduce risk through diversification. Diversification means investing in a variety of assets with different levels of risk and return, so that the overall risk of the portfolio is lower than the risk of any individual asset.

However, simply investing in a collection of assets is not enough. It is important to construct a portfolio that is optimized to achieve the best possible risk-return trade-off. A portfolio that is not optimized may have too much exposure to a particular asset or sector, which can increase risk and lead to lower returns.

Portfolio optimization provides a framework for constructing portfolios that are diversified and optimized to achieve the best possible risk-return trade-off. By taking into account the expected returns, variances, and covariances of the assets in a portfolio, investors can construct portfolios that offer the highest expected return for a given level of risk or the lowest possible risk for a given level of expected return. This can help investors to achieve their investment goals while minimizing risk.

Overall, portfolio optimization is a powerful tool that allows investors to construct diversified portfolios that are optimized to achieve their investment objectives while minimizing risk.

## 2. Types of Portfolio Optimization

### 2.1 Markowitz Portfolio Optimization

Markowitz portfolio optimization is a widely used investment strategy that aims to minimize the risk of a portfolio while maximizing its expected return. This strategy was first introduced by Harry Markowitz in 1952 and is based on the concept of diversification.

The basic idea behind Markowitz portfolio optimization is to select a combination of assets that have low correlations with each other. By doing this, the portfolio can achieve a higher level of diversification, which can help to reduce the risk of the portfolio without sacrificing returns. The portfolio optimization process involves two steps:

(1) calculating the expected returns and variances of the assets in the portfolio, and

(2) selecting the portfolio that minimizes the risk for a given level of expected return.

## 2.1.1 The steps involved are as follows

a. Data acquisition

For each of the six sectors that have been selected, the historical prices of the top ten stocks are extracted using the YahooFinancials function in the yahoofinancial library of Python. The ticker names of the stocks are passed as the parameters to the YahooFinancials function with the start_date parameter set to '2006-01-01', the end_date parameter set to '2020-12-31', and the time_interval parameter set to 'daily'. In this way, the stock prices are extracted from the Yahoo Finance website from 17 November 2017 to 27 February 2020. The stock data have the following attributes: open, high, low, close, volume, and adjusted_close. Since the current work is based on univariate analysis, the variable close is chosen as the variable of interest, and the remaining variables are not considered. The univariate close values of the five stocks for a given sector from 17 November 2017 to 30 August 2019 are used for training the portfolio models whereas for testing we have used closed prices from 03 September 2019 to 27 February 2020.

b. Computation of return and volatility

Using the training data for the ten stocks in each sector, the daily return and log return values of each stock of that sector are computed. The daily return values are the percentage changes.

in the daily close values over successive days. For computing the daily return, the pct_change function of Python is used. Based on the daily return values, the daily volatility, and the annual volatility of the ten stocks of each sector are computed. The daily volatility is defined as the standard deviation of the daily return values. The daily volatility, on multiplication by a factor of the square root of 252, yields the value of the annual volatility.

Here, there is a standard assumption of 252 working days in a year for a stock market. The annual volatility value of a stock quantifies the risk associated with stock from the point of view of an investor, as it indicates the amount of variability in its price. For computing the volatility of stocks, the std function of Python is used. The daily return values are also aggregated into annual return values for each stock for every sector.

c. Computation of covariance and correlation matrices

After computing the volatilities and return of the stocks, the covariance, and the correlation matrices for the ten stocks in each sector are computed using the records in the training dataset. These matrices help us in understanding the strength of association between a pair of stock prices in each sector. Any pair exhibiting a high value of correlation coefficient indicates a strong association between them. The Python functions cov and corr are used to compute the covariance and correlation matrices. A good portfolio aims to minimize the risk while optimizing the return. Risk minimization of a portfolio requires identifying stocks that have low correlation among themselves so that a higher diversity can be achieved. Hence, computation and analysis of the covariance and correlation matrices of the stocks are of critical importance.

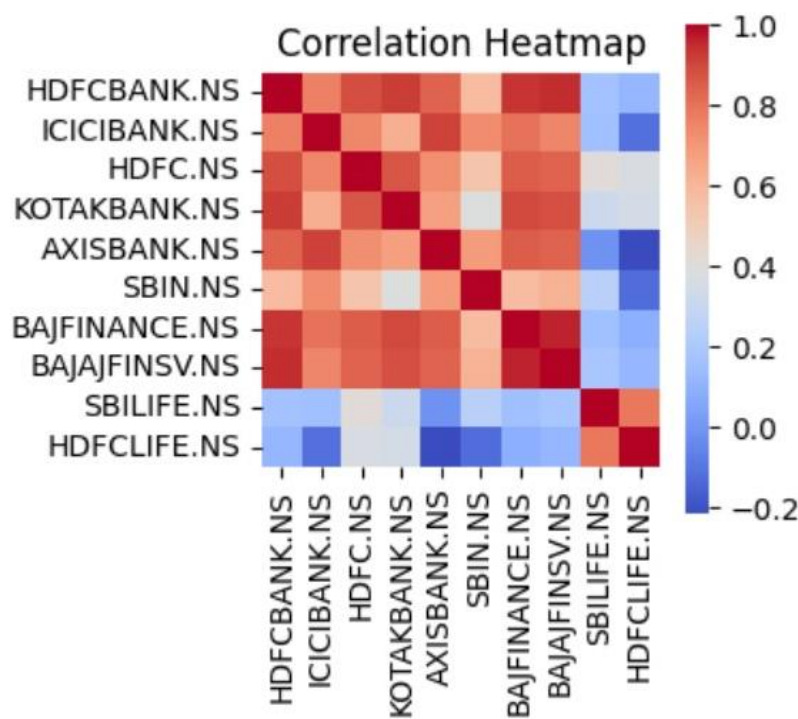

*Fig 1. Correlation Heatmap*

d. Computation of the expected return and risk of portfolios

At this step, a deeper analysis is done on the historical prices of the ten stocks in each of the seven sectors. First, for each sector, a portfolio is constructed using the ten stocks, with each stock carrying equal weight. Since there are ten stocks in a sector (i.e., in a portfolio), each stock is assigned a weight of 0.1. Based on the training dataset and using an equal-weight portfolio, the yearly return and risk (i.e., volatility) of each portfolio are computed. The computation of the expected return of a portfolio is done using (1). In (1), E(R) denotes the expected return of a portfolio consisting of n stocks, which are denoted as S1, S2, …Sn. The weights associated with the stocks are represented by $w_i$'s.

The yearly return and the yearly volatility of the equal-weight portfolio of each sector are computed using the training dataset. For this purpose, the mean of the yearly return values is derived using the resample function in Python with a parameter 'Y'. Yearly volatility values of the stocks in the equal-weight portfolio are derived by multiplying the daily volatility values by the square root of 250, assuming that there are, on average, 252 working days in a year for a stock exchange. The equal-weight portfolio of a sector gives one an idea about the overall profitability and risk associated with each sector over the training period. However, for future investments, their usefulness is very limited. Every stock in a portfolio does not contribute equally to its return and the risk. Hence, the computations of minimum risk and optimal risk portfolios are done in the next steps.

e. Building the minimum risk portfolio.

The minimum risk portfolio is built for each sector using the records in its training dataset. The minimum risk portfolio is characterized by its minimum variance. The variance of a portfolio is a metric computed using the variances of each stock in the portfolio as well as the covariances between each pair of stocks in the portfolio. The variance of a portfolio is computed using (2).

In (2), $w_i$ and $\sigma_i$ represent the weight associated with stock i and the standard deviation of the historical prices of the stock i. The covariance among the historical prices of stock i and stock j is denoted as $Cov(i, j)$. In the present work, there are ten stocks in a portfolio. Hence, 55 terms are involved in the computation of variance of each portfolio, 10 terms for the weighted variances, and 45 terms for the weighted covariances. For building the minimum risk portfolios, one needs to find the combination of $w_i$'s that minimizes the variance of the portfolio.

For finding the minimum risk portfolio, first, the efficient frontier for each portfolio is plotted. For a given portfolio of stocks, the efficient frontier is the contour with returns plotted along the y-axis and the volatility (i.e., risk) on the x-axis. The points of an efficient frontier denote the points with the maximum return for a given value of volatility or the minimum value of volatility for a given value of the return. Since, for an efficient frontier, the volatility is plotted along the x-axis, the minimum risk portfolio is identified by the leftmost point lying on the efficient frontier. For plotting the contour of the efficient frontier, the weights are assigned randomly to the five stocks in a portfolio in a loop and iterate the loop 10,100 times in a Python program. The iteration produces 10,000 points, each point representing a portfolio. The minimum risk portfolio is identified by detecting the leftmost point on the efficient frontier.

f. Computing the optimum risk portfolio.

The model calculates the optimal portfolio by solving the following quadratic optimization problem:

minimize: $w^T \Sigma w$

- subject to: $w^T \mu \geq r$
- $\Sigma w = 1$

where:

$w$ = vector of portfolio weights

$\mu$ = vector of expected returns of assets

$\Sigma$ = covariance matrix of asset returns

$r$ = target return

$\Sigma w$ = constraint that weights sum to one

The objective function represents the portfolio risk, which is the weighted sum of the variances and covariances of the assets in the portfolio. The first constraint ensures that the portfolio achieves a target expected return, while the second constraint ensures that the portfolio weights sum to one.

The investors in the stock markets are usually not interested in the minimum risk portfolios as the return values are usually low. In most cases, the investors are ready to incur some amount of risk if the associated return values are high. For computing the optimum risk portfolio, the metric Sharpe Ratio of a portfolio is used. The Sharpe Ratio of a portfolio is given by (3).

$$Sharpe\ Ratio = \frac{R_C - R_f}{\sigma_C} \qquad (3)$$

In (3), Rc, Rf, and σc denote the return of the current portfolio, the risk-free portfolio, and the standard deviation of the current portfolio, respectively. Here, the risk-free portfolio is a portfolio with a volatility value of 1%. The optimum-risk portfolio is the one that maximizes the Sharpe Ratio for a set of stocks. This portfolio makes an optimization between the return and the risk of a portfolio.

It yields a substantially higher return than the minimum risk portfolio, with a very nominal increase in the risk, and hence, maximizing the value of the Sharpe ratio. The optimal portfolio is identified using the idmax function in Python over the set of the Sharpe Ratio values computed for all the points of an efficient frontier. The optimal portfolio is the portfolio that lies on the efficient frontier and has the highest Sharpe ratio, which is the ratio of excess return to volatility.

```
Variance-Covariance Matrix:
[[ 7.59479836e+03  3.19610594e+03  1.14209316e+04  1.22017028e+04
   6.78675964e+03  1.56736408e+03  4.81267811e+04  8.20983565e+03
   9.42141405e+02  5.29942024e+02]
 [ 3.19610594e+03  2.30246869e+03  5.25192451e+03  4.58025931e+03
   4.05078816e+03  1.09376547e+03  2.27098002e+04  3.58345177e+03
   4.95392630e+02 -2.78485906e+02]
 [ 1.14209316e+04  5.25192451e+03  2.16545674e+04  1.94729926e+04
   9.88173861e+03  2.50631879e+03  7.40716867e+04  1.22523165e+04
   3.96573053e+03  2.76852525e+03]
 [ 1.22017028e+04  4.58025931e+03  1.94729926e+04  2.31512659e+04
   9.50868992e+03  1.83593015e+03  8.04669550e+04  1.33375519e+04
   3.17146666e+03  2.68556213e+03]
 [ 6.78675964e+03  4.05078816e+03  9.88173861e+03  9.50868992e+03
   8.53621897e+03  1.98097758e+03  4.64928134e+04  7.69887422e+03
  -6.00562287e+01 -1.02309653e+03]
 [ 1.56736408e+03  1.09376547e+03  2.50631879e+03  1.83593015e+03
   1.98097758e+03  9.66664674e+02  1.05468521e+04  1.90713397e+03
   5.05352571e+02 -1.95656078e+02]
 [ 4.81267811e+04  2.27098002e+04  7.40716867e+04  8.04669550e+04
   4.64928134e+04  1.05468521e+04  3.46525199e+05  5.64657074e+04
   6.00971139e+03  2.66899086e+03]
 [ 8.20983565e+03  3.58345177e+03  1.22523165e+04  1.33375519e+04
   7.69887422e+03  1.90713397e+03  5.64657074e+04  9.86145529e+03
   1.23533014e+03  5.92701335e+02]
 [ 9.42141405e+02  4.95392630e+02  3.96573053e+03  3.17146666e+03
  -6.00562287e+01  5.05352571e+02  6.00971139e+03  1.23533014e+03
   4.35984594e+03  2.66061679e+03]
 [ 5.29942024e+02 -2.78485906e+02  2.76852525e+03  2.68556213e+03
  -1.02309653e+03 -1.95656078e+02  2.66899086e+03  5.92701335e+02
   2.66061679e+03  2.60517092e+03]]
```

*Fig 2. Variance – Covariance Matrix*

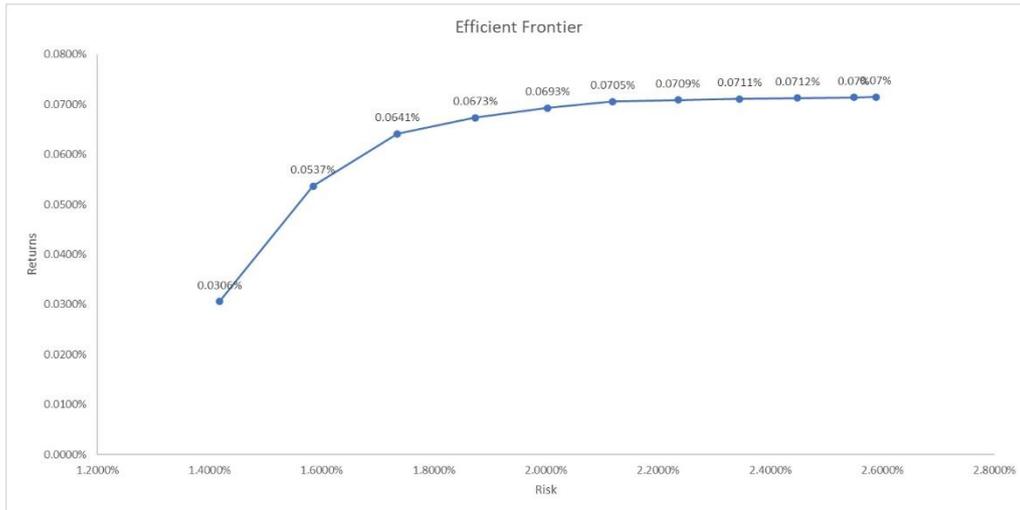

*Fig 3. Efficient Frontier*

## 2.1.2 Constraints and Limitations

The Markowitz Portfolio Optimization model makes several assumptions about the behaviour of financial markets and the characteristics of assets that are included in a portfolio. These assumptions are:

- Normal distribution of asset returns: The Markowitz model assumes that asset returns are normally distributed, which means that they follow a bell-shaped curve. This assumption is important because it allows the model to use statistical methods to estimate the expected return and variance of assets.
- Investors are risk-averse: The Markowitz model assumes that investors are risk-averse, which means that they prefer portfolios that provide a higher expected return for a given level of risk or a lower level of risk for a given expected return.
- Single period investment: The Markowitz model assumes that the investment period is a single period, which means that it does not consider the effects of compounding over time.
- No taxes, transaction costs, or other frictions: The Markowitz model assumes that there are no taxes, transaction costs, or other frictions associated with buying or selling assets.
- Stable correlation between assets: The Markowitz model assumes that the correlation between assets is stable over time. This assumption can be problematic in volatile markets where correlations may change rapidly.
- No short-selling restrictions: The Markowitz model assumes that investors can sell assets short, which means they can borrow assets and sell them with the expectation of buying them back at a lower price in the future. This assumption can be problematic in practice because short-selling is not always allowed or feasible.
- Expected returns are constant: The Markowitz model assumes that expected returns are constant over time. This assumption may not hold true in practice, as asset returns can be influenced by a variety of factors, including macroeconomic conditions, company performance, and changes in investor sentiment.

It is important to keep these assumptions in mind when applying the Markowitz Portfolio Optimization model in practice. Violations of these assumptions can lead to suboptimal portfolio performance and investors should consider the limitations of the model before making investment decisions.

## 2.2 Hierarchical Risk Parity Portfolio (HRP)

The HRP (Hierarchical Risk Parity) portfolio optimization method is a risk-based approach to asset allocation that aims to provide a balanced and diversified portfolio. The HRP method was developed by Marcos López de Prado in his paper "Building Diversified Portfolios that Outperform Out-of-Sample" in 2016.

The HRP method involves three main steps:

### 2.2.1 Hierarchical Clustering

Breaks down our assets into hierarchical clusters. It is used to place our assets into clusters. The objective is to build a hierarchical tree in which our assets are all clustered on different levels. Distance matrix is used as input for clustering. The similar assets are placed together as the distance between them are less. Ultimately at this stage clusters are formed using agglomerative hierarchical clustering method.

Given T x N matrix of stock returns, calculate the correlation of each stock's return with the other stocks giving us a NxN matrix of their correlation values (ρ).

The correlation matrix is then converted into a correlation distance matrix D such that D(i,j) indicates distance between the i th and j th asset where D(i,j) is given by,

Now, we calculate another distance matrix by,

$$D(i,j) = \sqrt{0.5 * (1-\rho(i,j))}$$

$$\overline{D}(i,j) = \sqrt{\sum_{k=1}^{N}(D(k,i)-D(k,j))^2}$$

It is formed by taking the Euclidean distance between all the columns in a pair-wise manner.

A quick explanation regarding the difference between $D$ and $\overline{D}$ for two assets i and j, $D(i,j)$ is the is the distance between the two assets while $\overline{D}(i,j)$ indicates the closeness in similarity of these assets with the rest of the portfolio. This becomes obvious when we look at the formula for calculating $\overline{D}$ – we sum over the squared difference of distances of i and j from the other stocks. Hence, a lower value means that assets i and j are similarly correlated with the other stocks in our portfolio.

We start forming clusters of assets using these distances in a recursive manner. Let us denote the set of clusters as $U$. The first cluster $(i^*, j^*)$ is calculated as,

$$U[1] = argmin_{(i,j)}\overline{D}(i,j)$$

We now update the distance matrix D by calculating the distances of other items from the newly formed cluster. This step is called linkage clustering and there are different ways of doing this. Hierarchical Risk Parity uses single linkage clustering which means the distances between two clusters is defined by a single element pair – those two elements which are closest to each other.

We remove the columns and rows corresponding to the new cluster – in this case we remove rows and columns for stocks, a and b where a and b are stocks that are clustered together previously. For calculating the distance of an asset outside this cluster, we use the following formula.

$$\overline{D}(i, U[1]) = min(\overline{D}(i, a), \overline{D}(i, b))$$

Using the above formula, we calculate distances for the rest of the assets or stocks from the cluster (a, b).

In this way, we go on recursively combining assets into clusters and updating the distance matrix until we are left with one giant cluster of stocks.

Finally, in hierarchical clustering, the clusters are always visualised in the form of a nice cluster diagram called *dendrogram*. Below is the image of the hierarchical clusters for our stock data.

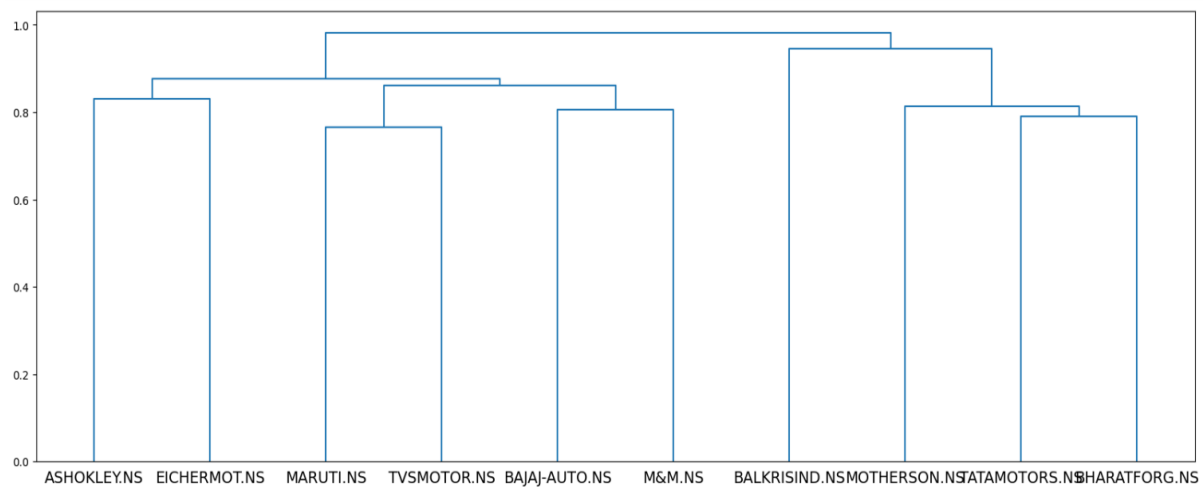

*Fig 4. Hierarchical Clustering Dendrogram for Auto Sector*

## 2.2.2 Quasi-Diagonalization

Quasi-diagonalization is like rearranging the data to show the inherent clusters clearly. Using the order of hierarchical clusters from the previous step, we rearrange the rows and columns of the covariance matrix of stocks so that similar investments are placed together and dissimilar investments are placed far apart. This rearranges the original covariance matrix of stocks so that larger covariances are placed along the diagonal and smaller ones around this diagonal and since the off-diagonal elements are not completely zero, this is called a *quasi-diagonal covariance matrix*.

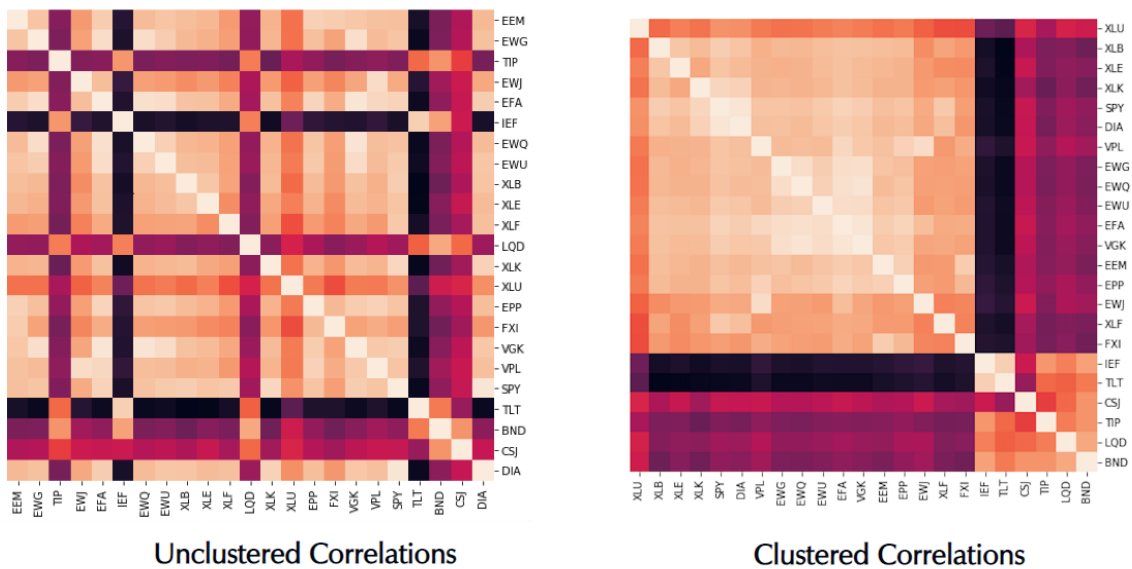

*Fig 5. Quasi-Diagonal Matrix*

We see that initially; the asset clusters are broken down into small sub-sections and after quasi-diagonalization the clustering structure becomes more evident.

### 2.2.3 Recursive Bisection

Weights are assigned to each of the assets and these weights are inversely proportional to the risk or standard deviation of that asset.

In this step, the actual portfolio weights are assigned to our assets in a top-down recursive manner. So, this step breaks each cluster into sub-clusters by starting with our largest cluster and moving down our tree in a top-down manner. Recursive bisection makes use of our quasi-diagonalized covariance matrix for recursing into the clusters under the assumption that for a diagonal matrix, the inverse-variance allocation is the most optimal allocation.

This is the final and the most important step of this algorithm where the actual weights are assigned to the assets in our portfolio.

We initialize the weights of the assets,

$$W_i = 1, \forall i = 1, \ldots, N$$

At the end of the tree-clustering step, we were left with one giant cluster with subsequent clusters nested within each other. We now break each cluster into two sub-clusters by starting with the topmost cluster and traversing in a top-down manner. This is where Hierarchical Risk Parity makes use of step-2 to quasi-diagonalize the covariance matrix and uses this new matrix for recursing into the clusters.

Hierarchical tree clustering forms a binary tree where each cluster has a left and right child cluster V1 and V2 are the corresponding variances. For each of these sub-clusters, we calculate its variance,

$$V_{adj} = w^T V w$$

where,

$$w = \frac{diag[V]^{-1}}{trace(diag[V]^{-1})}$$

A weighting factor is calculated based on the new covariance matrix

$$\alpha_1 = 1 - \frac{V_1}{V_1 + V_2}; \alpha_2 = 1 - \alpha_1$$

Once the weights are assigned to each cluster the same process is executed recursively until all the weights are assigned to the stocks within every cluster.

## 2.2.4 Advantages of HRP over Markowitz Model

HRP (Hierarchical Risk Parity) is a portfolio optimization method that has gained popularity in recent years due to its ability to provide a more stable allocation of assets and reduce the sensitivity of the portfolio to estimation errors.

- HRP is more robust to estimation errors in the covariance matrix. Markowitz model relies heavily on accurate estimates of the covariance matrix, which can be difficult to obtain, especially when dealing with a large number of assets. HRP, on the other hand, utilizes a hierarchical clustering algorithm that groups assets with similar risk profiles, reducing the reliance on precise covariance matrix estimates.
- HRP provides a more diversified portfolio. Markowitz model is known to produce portfolios that are heavily concentrated in a few assets. HRP, on the other hand, aims to balance risk across all assets in the portfolio, resulting in a more diversified portfolio.
- HRP is more stable over time. Markowitz model can be highly sensitive to changes in the market environment, which can result in significant changes to the portfolio allocation. HRP, on the other hand, provides a more stable allocation of assets that is less sensitive to short-term changes in the market.

### 2.2.5 Limitations of HRP

- HRP can be computationally intensive. The hierarchical clustering algorithm used in HRP can be time-consuming, especially when dealing with a large number of assets.
- HRP does not explicitly consider the expected returns of assets. The method focuses solely on balancing the risk across all assets in the portfolio, which may result in lower expected returns compared to other portfolio optimization methods.
- HRP assumes that asset returns follow a normal distribution. However, in practice, asset returns may not always follow a normal distribution, which may affect the accuracy of the method.
- HRP is a relatively new method, and its performance has not been extensively tested over long periods of time. Therefore, the method may be subject to unforeseen risks and limitations.

## 3. Model Comparisons (MVP & HRP)

In our study we have taken six different sectors of Indian economy i.e. Auto, Pharma, Banking, Finance, FMCG and IT. In each sector we have taken 10 most important stocks and applied MVP and HRP on stocks selected from every sector.

For evaluation we have taken data from yahoo financial from 17$^{th}$ Nov 2017 to 28$^{th}$ Feb 2020.

### AUTO SECTOR

The 10 most significant stocks from NSE -auto sector index are taken and they are as follows: Tata Motors, Ashoke Leyland, Eicher Motors ,Maruti Suzuki, TVS Motors, Motherson,  Bharat Forge, Bajaj-Auto, Balkrisind

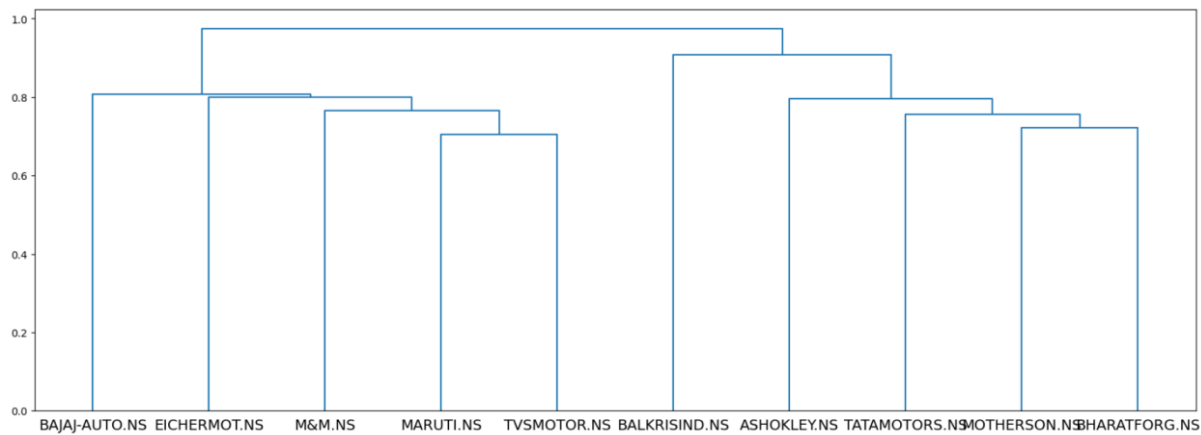

*Fig 6. The agglomerative clustering of the auto sector stocks – the dendrogram formed on the training data from 17-Nov-2017 to 30-Aug-2019.*

The dendrogram of the clustering of the stocks of the auto sector is shown in figure above. The *y*-axis of the dendrogram depicts the *ward linkage* values, where a longer length of the arms signifies a higher distance, and hence less compactness in the cluster formed.

[Type here]

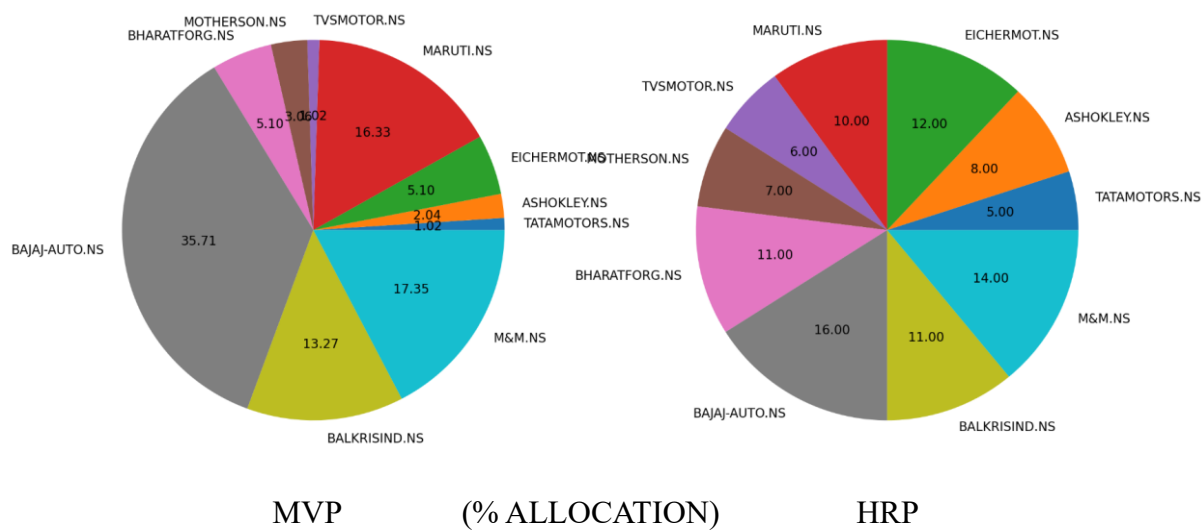

MVP　　　　　(% ALLOCATION)　　　　HRP

*Fig 7. The allocation of weights to the auto sector stocks by the MVP and the HRP portfolios based on stock price data from 17-Nov- 2017 to 30-Aug-2019.*

Fig 7 depicts the weight allocations by the MVP and the HRP portfolios for the auto sector. It is clear that HRP has tried to make the portfolio more diversified by more evenly allocating the weights to the stocks than the MVP portfolio.

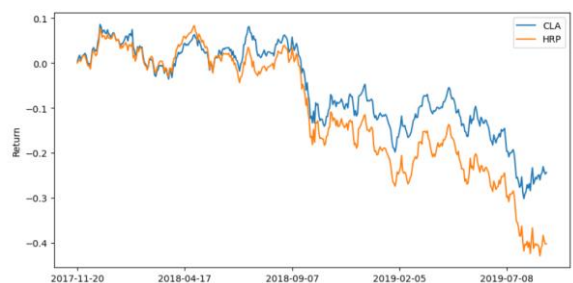　　　　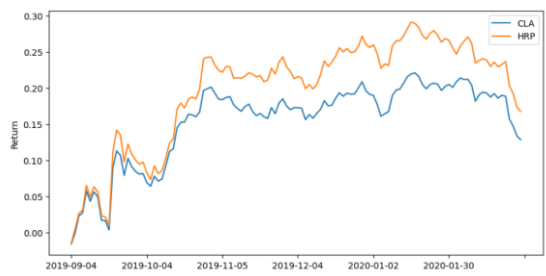

*Fig. 8.1.*　　　　　　　　　　　　　　*Fig. 8.2.*

*Fig. 8. The returns of the CLA and the HRP portfolios for the auto sector stocks.*

Fig 8.1 and Fig 8.2 show the results of back testing for the training data(17-Nov-2017 to 30-Aug-2019) and the test data(03-Sep-2019 to 27-Feb-2020) respectively.

| Portfolio | Training | | Testing | |
|---|---|---|---|---|
| | Risk | Sharpe Ratio | Risk | Sharpe Ratio |
| MVP | 0.184908 | -0.756289 | 0.216563 | 1.245047 |
| HRP | 0.200776 | -1.151904 | 0.256510 | 1.371209 |

*Table 1: Results of MVP and HRP over training and testing data*

The summary of the back testing results is presented in Table 1. While the HRP portfolio is found to have produced a lower Sharpe ratio *and* higher Risk than those of MVP Portfolio for the training (i.e., in-sample) data, the HRP portfolio has outperformed MVP portfolio on the test (i.e., out-of-sample) data producing a higher value of Sharpe ratio, with slightly higher volatility.

## PHARMA SECTOR

The 10 most significant stocks from NSE -Pharma sector index are taken and they are as follows: Torrent Pharma, Biocon, Dr Reddy Laboratory, Sun Pharma , Aurobindo Pharma, Syngene , Lal Path Lab, Alkem , Ipcalab , Laurus Labs.

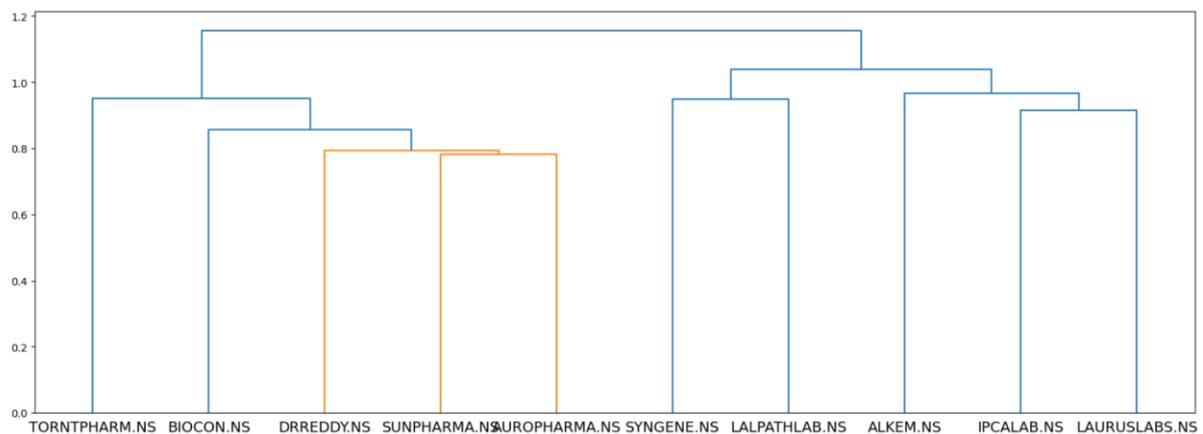

**Fig. 9.** *The agglomerative clustering of the Pharma sector stocks – the dendrogram formed on the training data from Nov-17-2017 to 30-Aug-2019.*

The dendrogram of the clustering of the stocks of the auto sector is shown in Fig 9.. The *y*-axis of the dendrogram depicts the *ward linkage* values, where a longer length of the arms signifies a higher distance, and hence less compactness in the cluster formed.

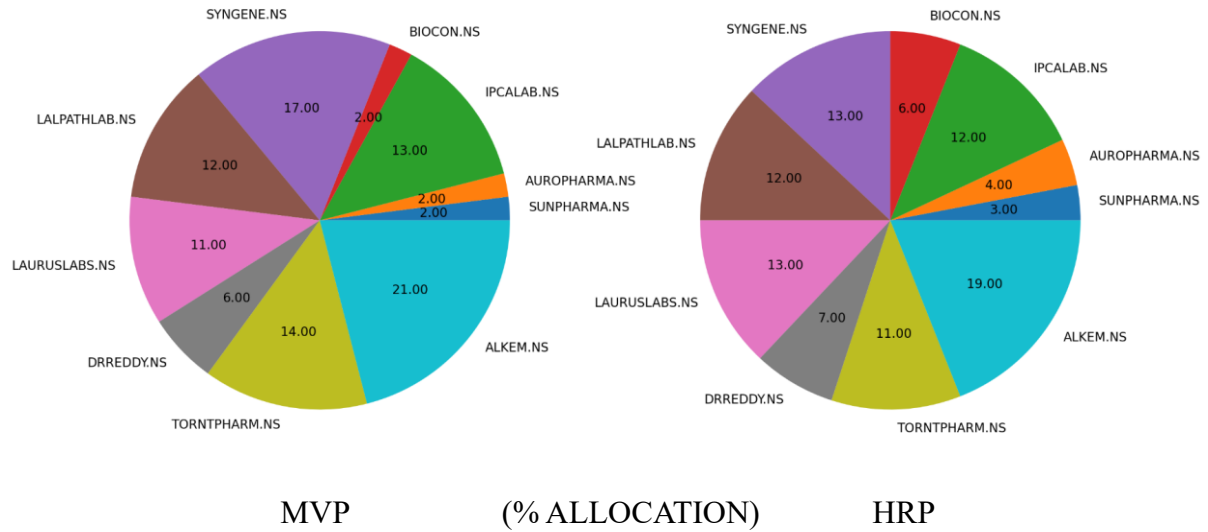

           MVP           (% ALLOCATION)         HRP

**Fig. 10.** The allocation of weights to the Pharma sector stocks by the MVP and the HRP portfolios based on stock price data from 17-Nov-2017 to 30-Aug-2019.

Fig 10 depicts the weight allocations by the MVP and the HRP portfolios for the Pharma sector. It is clear that HRP has tried to make the portfolio more diversified by more evenly allocating the weights to the stocks than the MVP portfolio.

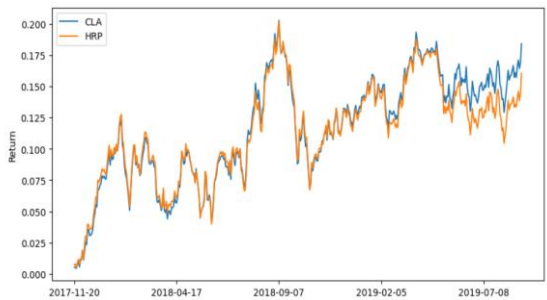 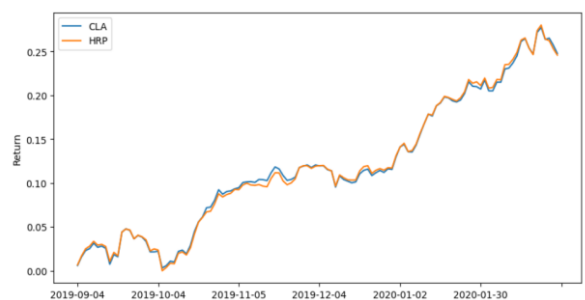

*Fig.11.1*                                         *Fig. 11.2*

**Fig. 11.** *The returns of the MVP and the HRP portfolios for the Pharma sector stocks.*

Fig 11.1 and Fig 11.2 show the results of back testing for the training data (17-Nov-2017 to 30-Aug-2019) and the test data(03-Sep-2019 to 27-Feb-2020) respectively.

|  | Training | | Testing | |
|---|---|---|---|---|
| Portfolio | Risk | Sharpe Ratio | Risk | Sharpe Ratio |
| MVP | 0.120684 | 0.877575 | 0.124738 | 4.172059 |
| HRP | 0.123018 | 0.750673 | 0.126812 | 4.071148 |

*Table: 2*

The summary of the backtesting results is presented in Table 2. While the MVP portfolio is found to have produced a Higher Sharpe ratio than those of HRP Portfolio for both training (i.e., in-sample) data and testing data(i.e., out-of-sample).

## IT SECTOR

The 10 most significant stocks from NSE- IT sector index are taken and they are as follows: Wipro, TechM, HCL Tech, TCS, INFY, COFORGE, LTIM, LTTS, MPHASIS, PERSISTENT.

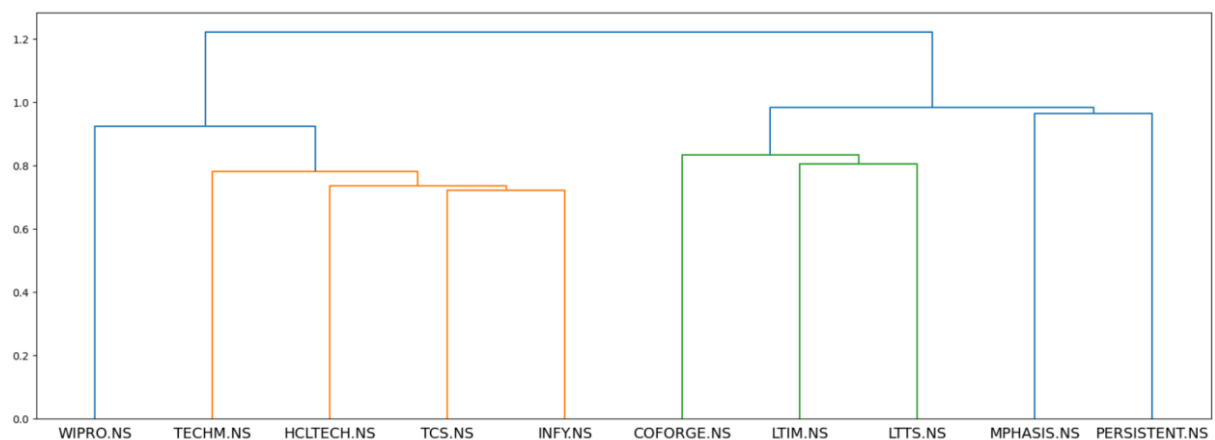

**Fig. 12.** *The agglomerative clustering of the IT sector stocks – the dendrogram formed on the training data from Nov17 2017 to 30-Aug-2019.*

The dendrogram of the clustering of the stocks of the auto sector is shown in Fig 12. The *y*-axis of the dendrogram depicts the *ward linkage* values, where a longer length of the arms signifies a higher distance, and hence less compactness in the cluster formed.

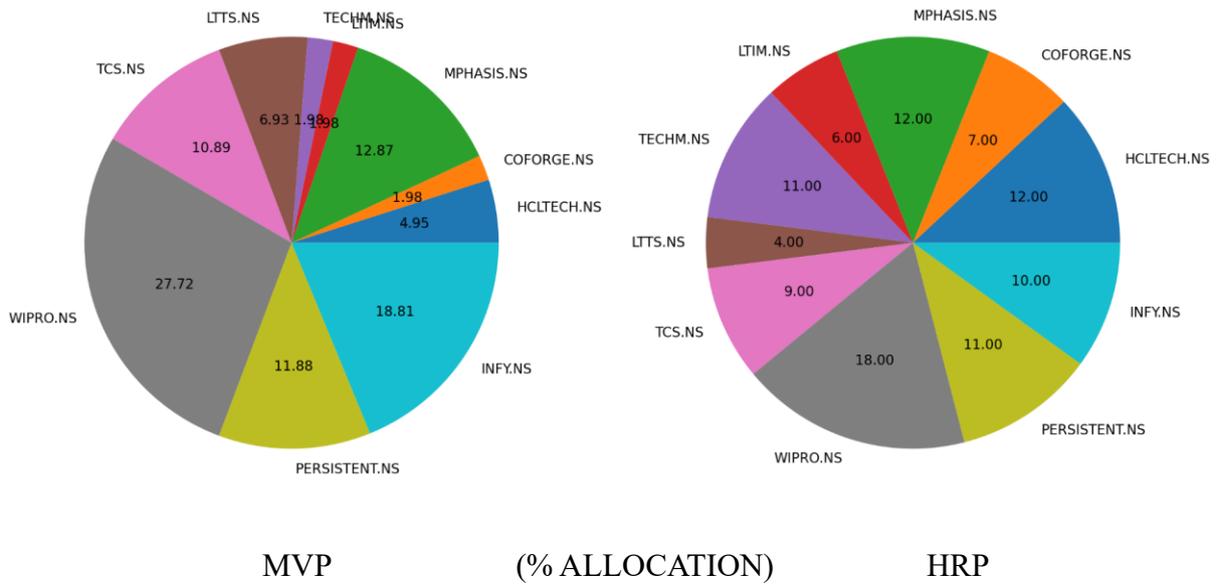

MVP          (% ALLOCATION)          HRP

***Fig.13.*** *The allocation of weights to the IT sector stocks by the MVP and the HRP portfolios based on stock price data from 17-Nov- 2017 to 30-Aug-2019.*

Fig 13 depicts the weight allocations by the MVP and the HRP portfolios for the auto sector. It is clear that HRP has tried to make the portfolio more Diversified by more evenly allocating the weights to the stocks than the MVP portfolio.

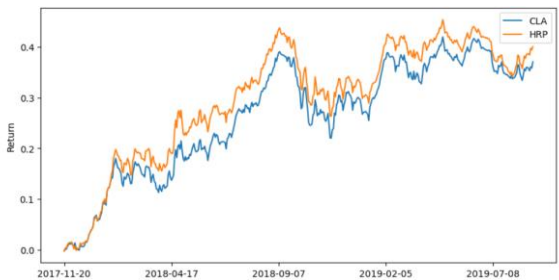 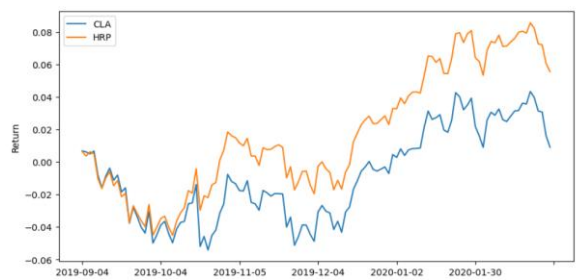

*Fig. 14.1*                        *Fig. 14.2*

***Fig.14.*** *The returns of the MVP and the HRP portfolios for the IT sector stocks.*

Fig 14.1 and Fig 14.2 show the results of back testing for the training data (17-Nov-2017 to 30-Aug-2019) and the test data(03-Sep-2019 to 27-Feb-2020) respectively.

| Portfolio | Training | | Testing | |
|---|---|---|---|---|
| | Risk | Sharpe Ratio | Risk | Sharpe Ratio |
| MVP | 0.151003 | 1.412311 | 0.137171 | 0.13802 |
| HRP | 0.156862 | 1.469323 | 0.124181 | 0.93751 |

*Table 3*

The summary of the backtesting results is presented in Table 3. The HRP portfolio is found to have produced a Higher Sharpe ratio than those of MVPPortfolio for both training (i.e., in-sample) data and testing data(i.e., out-of-sample).

## BANKING SECTOR

The 10 most significant stocks from NSE Banking sector index are taken and they are as follows: Federal Bank, IDFC First Bank, ICICI Bank, SBI , PNB, BANK OF BARODA, AU BANK, INDUSIND Bank, HDFC Bank, KOTAK Bank.

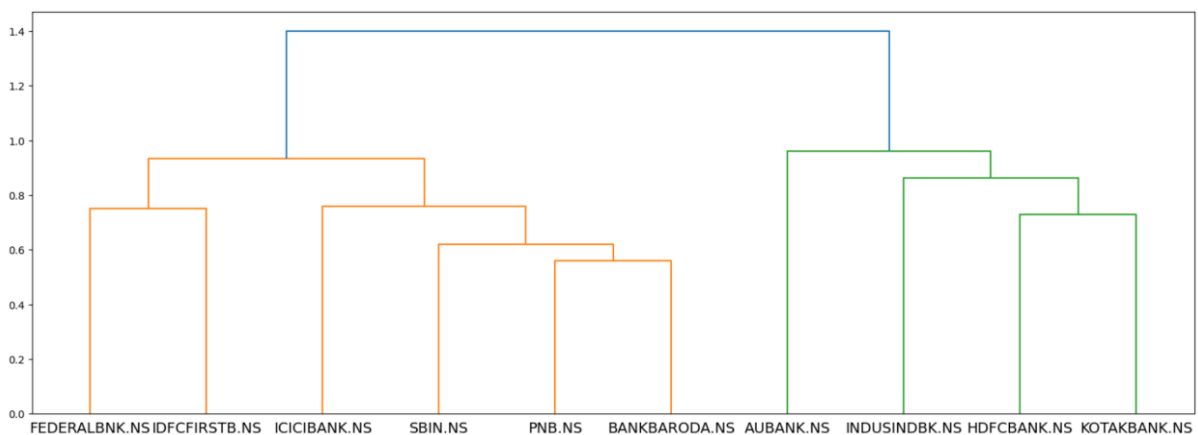

**Fig. 15.** *The agglomerative clustering of the IT sector stocks – the dendrogram formed on the training data from Nov17 2017 to 30-Aug-2019.*

The dendrogram of the clustering of the stocks of the auto sector is shown in Fig 15. The *y*-axis of the dendrogram depicts the *ward linkage* values, where a longer length of the arms signifies a higher distance, and hence less compactness in the cluster formed.

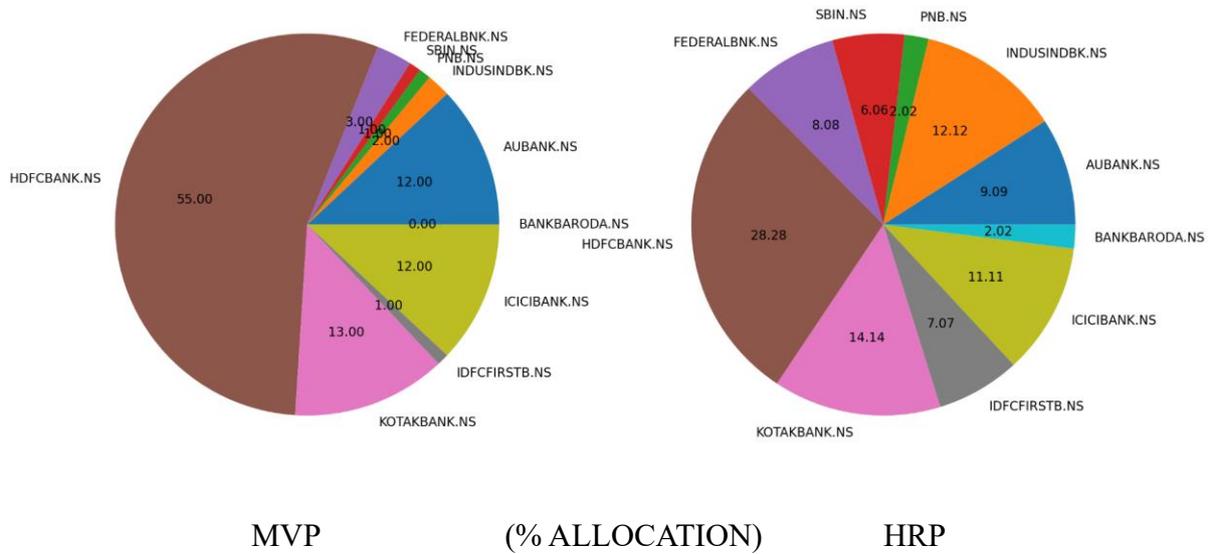

          MVP          (% ALLOCATION)          HRP

**Fig. 16.** *The allocation of weights to the Banking sector stocks by the MVP and the HRP portfolios based on stock price data from 17-Nov- 2017 to 30-Aug-2019.*

Fig 16 depicts the weight allocations by the MVP and the HRP portfolios for the auto sector. HRP has tried to make the portfolio more diversified by more evenly allocating the weights to the stocks than the MVP portfolio.

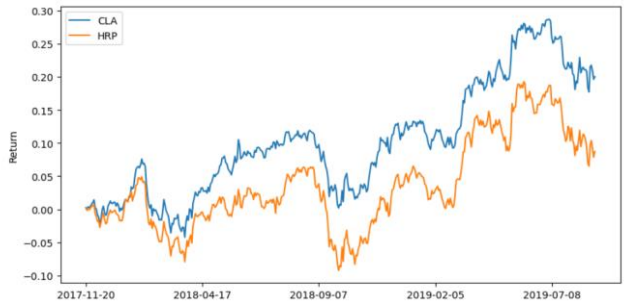 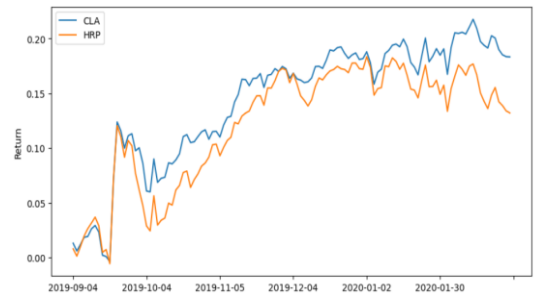

          *Fig. 17.1*                                                *Fig. 17.2*

**Fig. 17.** *The returns of the MVP and the HRP portfolios for the Banking sector stocks.*

Fig 17.1 and Fig 17.2 shows the results of backtesting for the training data (17-Nov-2017 to 30-Aug-2019) and the test data(03-Sep-2019 to 27-Feb-2020) respectively.

| Portfolio | Training | | Testing | |
|---|---|---|---|---|
| | Risk | Sharpe Ratio | Risk | Sharpe Ratio |
| MVP | 0.143586 | 0.801550 | 0.197891 | 1.945700 |
| HRP | 0.158120 | 0.317477 | 0.213296 | 1.300964 |

*Table 4*

The summary of the backtesting results is presented in Table 4. The MVP portfolio is found to have produced a higher Sharpe ratio than that of the HRP Portfolio for the training (i.e., in-sample) and testing (i.e., out-of-sample) data.

## FINANCE SECTOR

The 10 most significant stocks from the NSE-FINANCE sector index are taken and they are as follows: L&T FH, IBULHSGFIN, JFINANCIAL LICHSGFIN, M&MFIN, MOTILALOFS, TATAINVEST, BAJFINANCE, ISEC, EDELWEISS.

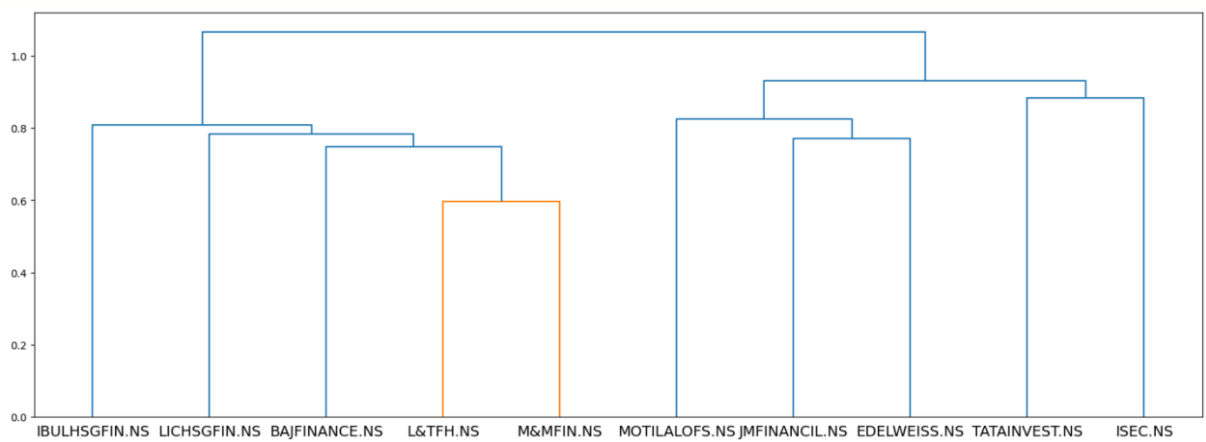

**Fig. 18.** *The agglomerative clustering of the financial sector stocks – the dendrogram formed on the training data from Nov 17 2017 to 30-Aug-2019.*

The dendrogram of the clustering of the stocks of the financial sector is shown in Fig 18. The *y*-axis of the dendrogram depicts the *ward linkage* values, where a longer length of the arms signifies a higher distance, and hence less compactness in the cluster formed.

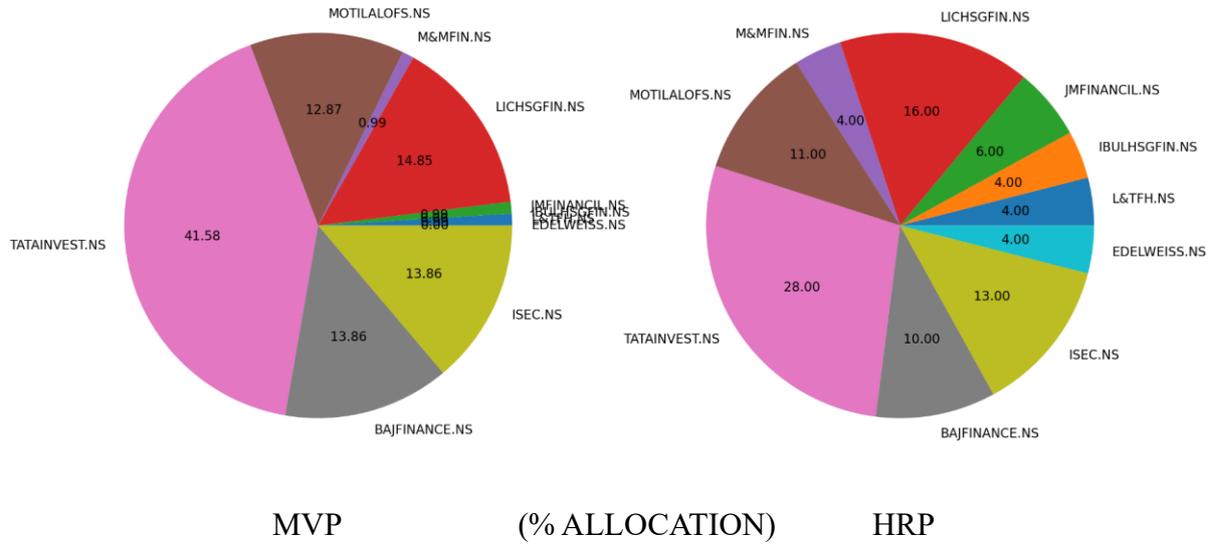

MVP            (% ALLOCATION)         HRP

***Fig. 19.*** *The allocation of weights to the Financial sector stocks by the MVP and the HRP portfolios based on stock price data from 17-Nov-2017 to 30-Aug-2019.*

Fig 19 depicts the weight allocations by the MVP and the HRP portfolios for the Financial sector. HRP has tried to make the portfolio more Diversified by more evenly allocating the weights to the stocks than the MVP portfolio.

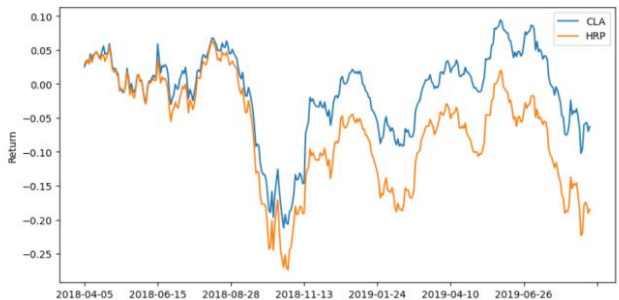 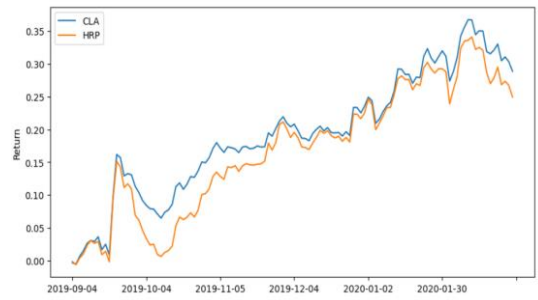

*Fig. 20.1.*                         *Fig. 20.2.*

***Fig.20.*** *The returns of the MVP and the HRP portfolios for the financial sector stocks.*

Fig 20.1 and Fig 20.2 shows the results of backtesting for the training data (17-Nov-2017 to 30-Aug-2019) and the test data(03-Sep-2019 to 27-Feb-2020) respectively.

|  | Training | | Testing | |
|---|---|---|---|---|
| Portfolio | Risk | Sharpe Ratio | Risk | Sharpe Ratio |
| MVP | 0.201811 | -0.22511 | 0.256589 | 2.363913 |
| HRP | 0.215537 | -0.62437 | 0.283526 | 1.847283 |

*Table 5*

The summary of the backtesting results is presented in Table 4. The MVP portfolio is found to have produced a higher Sharpe ratio than that of the HRP Portfolio for the training (i.e., in-sample) and testing (i.e., out-of-sample) data.

## FMCG SECTOR

The 10 most significant stocks from the NSE IT sector index are taken and they are as follows: BRITANIA, NESTLE, COLPAL, GODREJ, DABUR, HINDUSTAN UNILEVER, UBL, MCDOWELL.

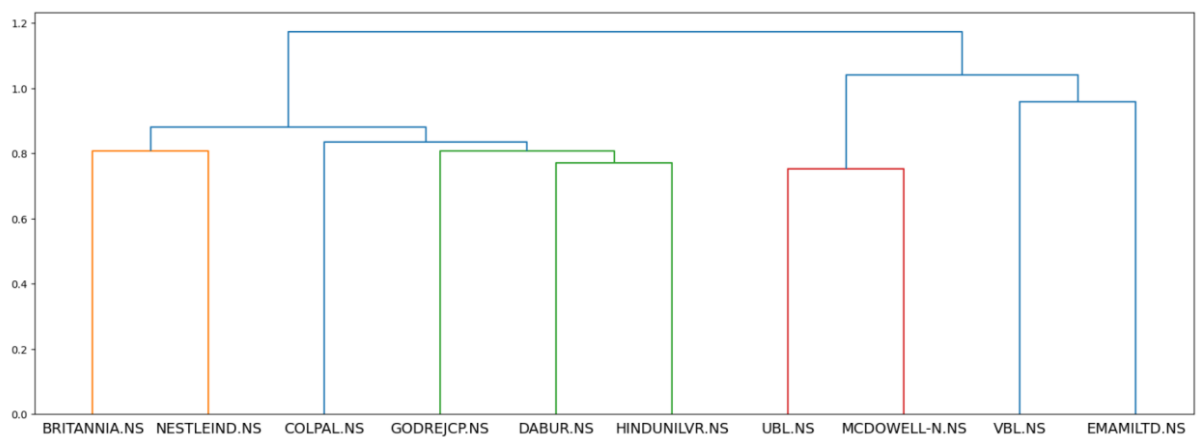

***Fig. 21.*** *The agglomerative clustering of the FMCG sector stocks – the dendrogram formed on the training data from Nov-1-2017 to 30-Aug-2019.*

The dendrogram of the clustering of the stocks of the FMCG sector is shown in Fig 21. The *y*-axis of the dendrogram depicts the *ward linkage* values, where a longer length of the arms signifies a higher distance, and hence less compactness in the cluster formed.

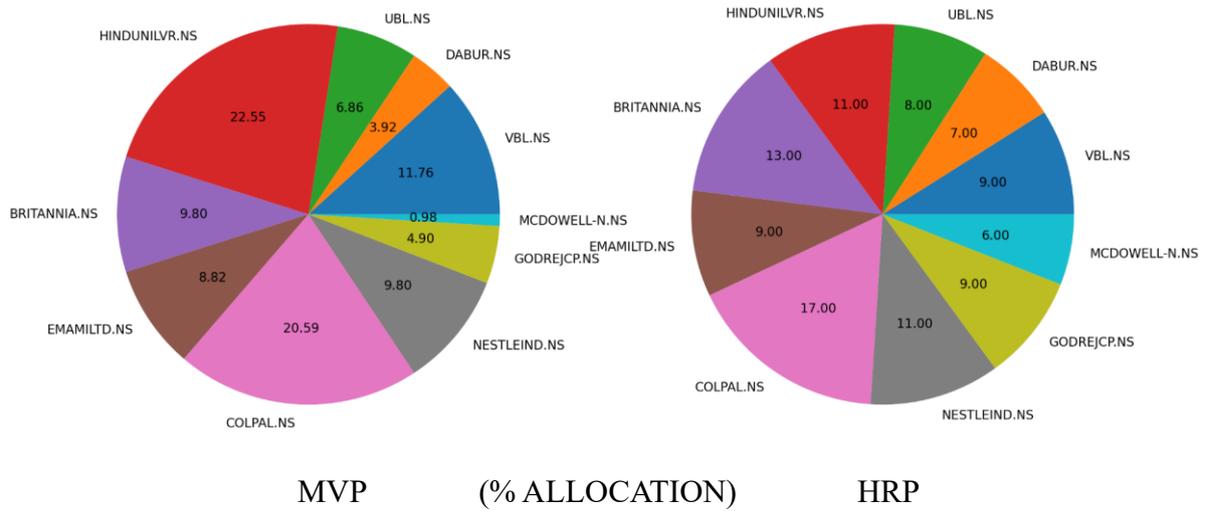

MVP             (% ALLOCATION)             HRP

***Fig. 22.*** *The allocation of weights to the FMCG sector stocks by the MVP and the HRP portfolios based on stock price data from 17-Nov-2017 to 30-Aug-2019.*

Fig 22 depicts the weight allocations by the MVP and the HRP portfolios for the financial sector. HRP has tried to make the portfolio more Diversified by more evenly allocating the weights to the stocks than the MVP portfolio.

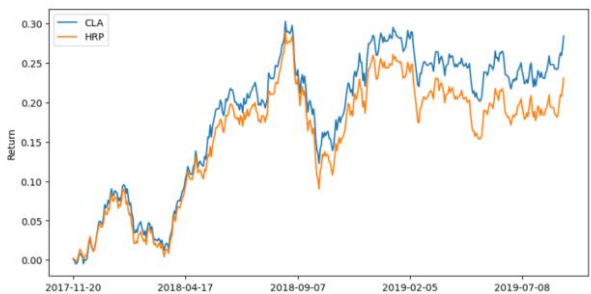 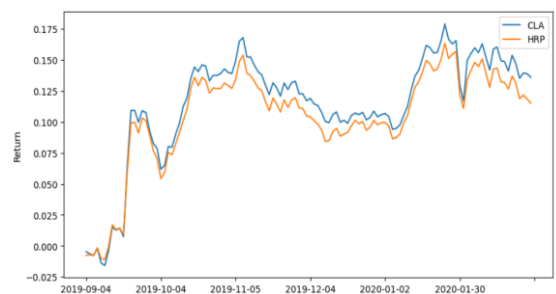

*Fig. 23.1.*                                    *Fig. 23.2.*

***Fig. 23.*** *The returns of the MVP and the HRP portfolios for the FMCG sector stocks.*

Fig 23.1 and Fig 23.2 shows the results of backtesting for the training data (17-Nov-2017 to 30-Aug-2019) and the test data(03-Sep-2019 to 27-Feb-2020) respectively.

| Portfolio | Training | | Testing | |
|---|---|---|---|---|
| | Risk | Sharpe Ratio | Risk | Sharpe Ratio |
| MVP | 0.133401 | 1.226017 | 0.134820 | 0.985869 |
| HRP | 0.174900 | 1.633208 | 0.161621 | 1.495954 |

*Table 6*

The summary of the backtesting results is presented in Table 4. The HRP portfolio is found to have produced a higher Sharpe ratio than that of MVP Portfolio for the training (i.e., in-sample) and testing (i.e., out-of-sample) data.

# 4. Reinforcement Learning

## 4.1 Introduction

Reinforcement learning: one of the most popular machine learning methods used today; enables a computer system to learn how to make choices by being rewarded for its successes. RL can be considered to be an extremely powerful tool for optimization and decision-making. It is an approach to machine learning in which the agents are trained to make a sequence of decisions. An agent can learn a sequence of actions to maximize rewards and thereby achieve the desired goal.

In portfolio optimization, the agent is the portfolio manager, the environment is the financial market, the actions are the portfolio allocations, and the rewards are the portfolio returns.

## 4.2 Need for RL in Portfolio Optimization

Portfolio optimization with reinforcement learning (RL) has several advantages over traditional portfolio optimization techniques. Here are some reasons why we need portfolio optimization with reinforcement learning:

- RL can handle complex and dynamic environments: The stock market is a complex and dynamic environment, and traditional portfolio optimization techniques may not be able to capture all the complexities and dynamics of the market. RL algorithms, on the other hand, are designed to work in complex and dynamic environments and can learn and adapt to changing market conditions.
- RL can learn from experience: Traditional portfolio optimization techniques rely on assumptions and statistical models to estimate expected returns, variances, and covariances of assets. RL algorithms, on the other hand, can learn from experience by interacting with the market and making investment decisions based on real-time market data.
- RL can optimize for non-linear objectives: Traditional portfolio optimization techniques are designed to optimize for linear objectives, such as expected return and variance. RL algorithms, on the other hand, can optimize for non-linear objectives, such as maximizing long-term growth or minimizing drawdowns.
- RL can handle large amounts of data: The stock market generates large amounts of data, including historical prices, news articles, and social media posts. RL algorithms can handle large amounts of data and extract useful patterns and insights that can inform investment decisions.

Overall, portfolio optimization with reinforcement learning has the potential to provide more accurate and effective investment strategies than traditional portfolio optimization techniques, especially in complex and dynamic markets. By leveraging the power of machine learning and artificial intelligence, investors can construct portfolios that are optimized to achieve their investment objectives while minimizing risk.

## 4.3 Terminologies

**Agent** – It is the sole decision-maker and learner. An agent is the entity that interacts with the environment and makes decisions based on the received information. The goal of the agent is to learn a policy or a strategy that maximizes the Sharpe ratio over time.

The agent observes the current state of the environment, selects an action based on its policy, and then receives a reward from the environment in response to its action. The agent uses this reward signal to update its policy and improve its decision-making over time.

**Environment** –It is a physical world where an agent learns and decides the actions to be performed. the environment is the external world or the system with which the agent interacts. The environment here is the Indian Stock Market.

The environment in RL is characterized by a set of states, actions, and rewards. At each time step, the agent receives a state from the environment, selects an action, and receives a reward in response to its action. The state represents the current situation of the environment, while the action represents the decision made by the agent based on the current state. The reward represents the feedback from the environment in response to the agent's action. The environment can be stochastic as the outcomes of actions are probabilistic. The RL algorithm learns to make better decisions by interacting with the environment and observing the resulting rewards.

**Action** – an action is a decision made by the agent based on the current state of the environment. The agent selects an action from a set of available actions based on its policy, which is a strategy for selecting actions in different states. The action taken by the agent modifies the state of the environment, which then generates a reward signal that the agent uses to update its policy. We have three actions: Buy, sell, or hold for adjusting the stock weights in a portfolio.

**State** – It is the current situation of the agent in the environment. The state can be thought of as the "input" to the agent, which it uses to decide on what action to take. The state may include information such as the location of the agent, the positions of other objects in the environment, the time of day, or any other relevant information that describes the environment at a given time. Here, the state is the Correlation Matrix of the stocks based on a specific time window.

**Reward** – a reward is a scalar feedback signal that the agent receives from the environment in response to its action. The reward reflects the quality of the agent's action in the given state of the environment. We have used the Sharpe ratio as the reward.

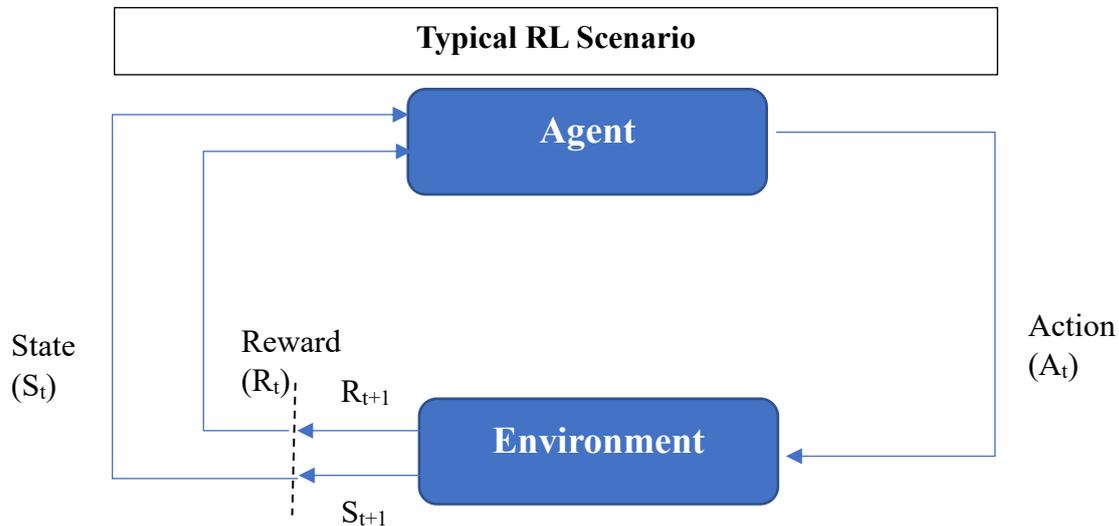

*Fig 24. Reinforcement Learning explained using a simple diagram*

In the diagram above, an agent performs an action in a particular environment and a particular state. A response is sent from the agent in the form of a reward and the new status information. State changes occur because of actions by the agent.

We have used Q-learning as the learning model for the Reinforcement learning environment.

**Value Function** – The reward concept is explained through the value function. This gives us the potential immediate reward along with all the potential future rewards. This value has to be discounted because future rewards inherently have uncertainty associated with them in the sense they may not be realized. So, we have a discounting factor of γ for this purpose.

$$Gt = R_{t+1} + \gamma R_{t+2} + \ldots = \sum \gamma k R_{t+k+1}$$

The state-value function, denoted by V(s), estimates the expected cumulative reward that the agent can achieve from a given state s, by following a given policy. It can be expressed as:

$$V(s) = E\,[\,R(t+1) + \gamma R(t+2) + \gamma\!\wedge\!2 R(t+3) + \ldots |\, S(t) = s\,]$$

where R(t+1), R(t+2), R(t+3), ... are the rewards received by the agent at time t+1, t+2, t+3, ... and γ is a discount factor that determines the importance of immediate versus future rewards.

The action-value function, denoted by Q(s,a), estimates the expected cumulative reward that the agent can achieve from a given state s and taking a given action a, by following a given policy. It can be expressed as:

$$Q(s,a) = E\,[\,R(t+1) + \gamma R(t+2) + \gamma^2 R(t+3) + \ldots \,|\, S(t) = s, A(t) = a\,]$$

where A(t) is the action taken by the agent at time t.

Both the state-value function and the action-value function are estimated using temporal difference learning. This method uses the observed rewards and transitions in the environment to update the estimates of the value function.

**Value Function and Q value interaction-** The product of probability distribution of action, state pair, and the q value associated with that state and action summed over all pairs gives us back the value function. This is intuitive as product of the probability of taking an action multiplied by the Q value associated with taking that action at that particular state and then summed up over all possible states should always give us the net expected value for being in some particular state.

$$\boldsymbol{V(s)} = \sum a \in A\, \boldsymbol{Q(s,a)}\pi(\boldsymbol{a|s})$$

Ultimately all of the above are used to optimize the Bellman Equation which is ultimately what we use to improve our algorithm iteratively

**Bellman Equation-** The Bellman equation is a fundamental concept in reinforcement learning (RL), which expresses the relationship between the value function of a state and the value function of its successor states. The Bellman equation for the state-value function can be expressed as follows:

$$V(s) = E\,[R(t+1) + \gamma V(S(t+1))\,|\,S(t) = s\,]$$

where V(s) is the value function of state s, R(t+1) is the immediate reward received after taking action A(t) in state S(t), S(t+1) is the successor state resulting from taking action A(t) in state S(t), and γ is a discount factor that determines the importance of immediate versus future rewards. This equation states that the value of a state is equal to the expected sum of the immediate reward and the discounted value of the next state. The discount factor γ determines the trade-off between immediate and future rewards, with values closer to 0 giving more weight to immediate rewards, and values closer to 1 giving more weight to future rewards

The Bellman equation for the Q-value function can be expressed as follows:

$$Q(s, a) = E\left[R(t+1) + \gamma max(Q(S(t+1), a')) \mid S(t) = s, A(t) = a\right]$$

where Q(s, a) is the Q-value function of state s and action a, max(Q(S(t+1), a')) is the maximum Q-value of the successor state S(t+1) and all possible actions a', R(t+1) is the immediate reward received after taking action A(t) in state S(t), and γ is a discount factor that determines the importance of immediate versus future rewards.

The Bellman equation plays a crucial role as it allows us to update the value function of a state based on the values of its successor states, and it helps to estimate the expected cumulative reward that an agent can achieve by following a given policy.

**Temporal Difference Learning-** Temporal difference (TD) learning is a type of reinforcement learning (RL) method that updates the value function estimates based on the temporal difference between the predicted and actual values. In TD learning, the agent interacts with the environment by taking actions and observing the resulting rewards and states. It uses the Bellman equation to update its estimates of the state-value or action-value function at each time step, by computing the difference between the predicted value and the actual reward received. The TD error is defined as the difference between the predicted value V(s) and the observed reward r(t+1) plus the discounted value of the next state V(S(t+1)):

$$\delta(t) = r(t+1) + \gamma V(S(t+1)) - V(S(t))$$

The value function estimate is then updated based on the TD error using a learning rate α:

$$V(S(t)) = V(S(t)) + \alpha\,\delta(t)$$

**Deep Q Learning-** Deep Q Learning is a type of reinforcement learning (RL) algorithm that uses deep neural networks to approximate the action-value function in RL problems with large and continuous state spaces. It is an extension of the Q-learning algorithm that can handle high-dimensional input data, such as images, by using a deep convolutional neural network (CNN) as a function approximator.

In Deep Q Learning, the agent interacts with the environment by taking actions and observing the resulting rewards and states. It uses a neural network to approximate the Q-function, which maps state-action pairs to their corresponding values. The network takes the state as input and outputs a Q-value for each possible action. The agent then selects the action with the highest Q-value.

During training, the network parameters are updated to minimize the mean squared error between the predicted Q-values and the target Q-values. The target Q-values are computed using the Bellman equation, with the maximum Q-value over all possible actions in the next state:

$$Q\_target = r + \gamma \max\_a' Q(s', a')$$

where r is the immediate reward received, γ is the discount factor, s' is the next state, a' is the next action, and Q(s', a') is the Q-value for the next state-action pair.

The loss function used in the training of the network is:

$$L = (Q\_target - Q(s, a)^2)$$

where Q(s, a) is the predicted Q-value for the current state-action pair.

**Equal Weighted Portfolio**-An equal-weighted portfolio is a type of investment portfolio in which each stock or security is given equal weight or allocation. This means that each security in the portfolio carries the same amount of importance or influence in determining the overall performance of the portfolio. For example, if an equal-weighted portfolio consists of ten stocks, each stock would make up 10% of the portfolio's total value. This differs from a market-cap-weighted portfolio, where each security's weight is based on its market capitalization, meaning that larger companies would have a higher weighting in the portfolio.

Equal-weighted portfolios can offer a more diversified exposure to the market and can potentially outperform market-cap-weighted portfolios in certain market conditions. However, they can also be riskier since smaller companies may have a larger impact on the portfolio's performance.

Since we took 10 stocks for analysis, each stock is weighting 0.10.

## 4.4 Program Design

1. Loading necessary python packages.

2. Loading Dataset (Extracted file from yahoo finance using python.

3. Data Preparation through checking for null values and replacing them with the previous values. Train Test Split (Train data: Nov 17 2017 to 30-Aug-2019) and Testing data(03-Sep-2019-27-Feb-2020). We have taken closed prices for our evaluation.

4. Model Development & Comparison with Equal Weighted Portfolio.

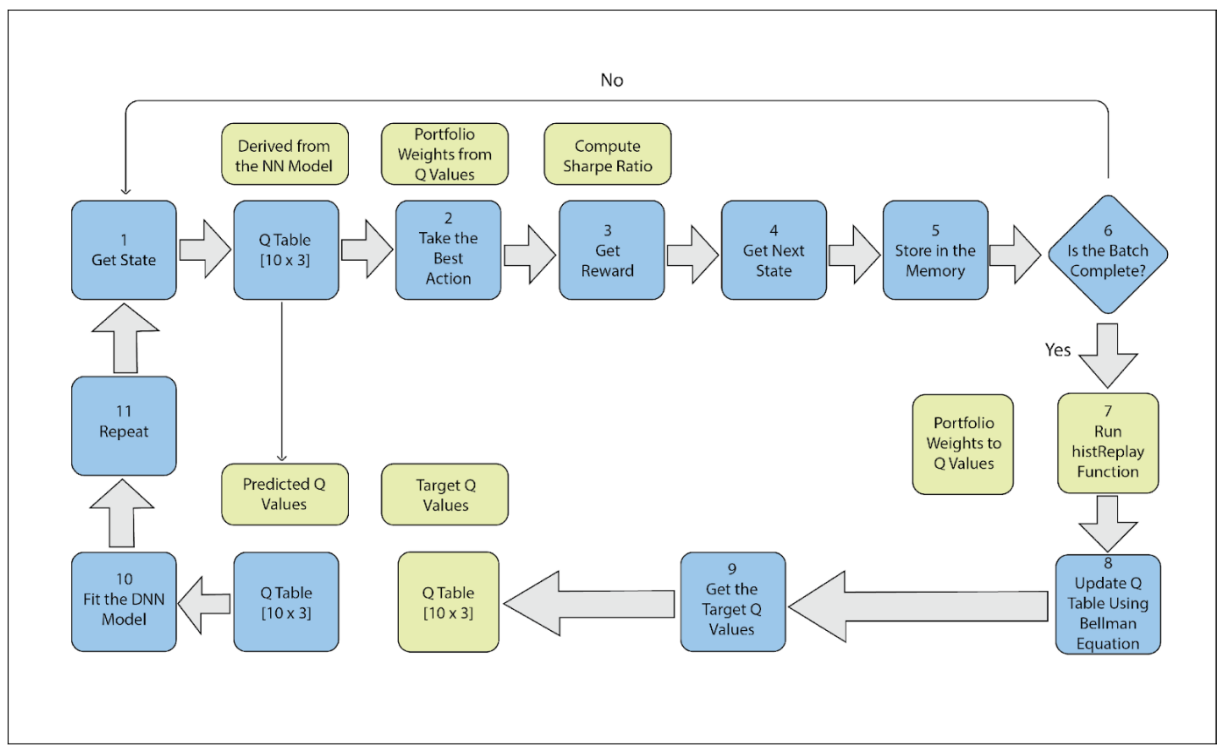

*Fig 25. The diagrammatic representation of the working of the RL program.*

## 4.5 Results & Findings

We have taken 6 sectors and applied RL for Portfolio optimization. Also we have benchmarked the performance of RL with that of an equal weight Portfolio.

For training and testing, we have used the same timeline i.e Training data (Nov 17 2017 to 30 Aug 2019) and Testing data (03 Sep 2019 to 27 Feb 2020).

Also, we have done hyperparameter tuning for a fixed set of values of the following hyperparameters: (Window size, Number of Episodes, batch size, Rebalance Period.)

Some other hyperparameters like learning rate, discount factor, and epsilon decay were kept constant.

The sectoral Performance in Testing is as follows:

## AUTO SECTOR

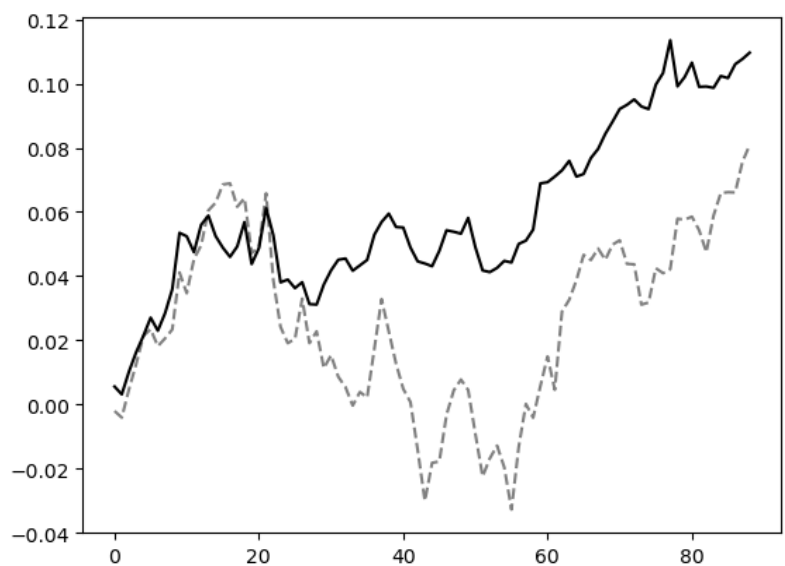

*Fig. 26. Solid line refers to the performance of the RL portfolio and the dotted line refers to the performance of Equal weight portfolio for the testing period(03-Sep-2019 to 27-Feb-2020).*

In the y-axis the cumulative returns are plotted whereas the x-axis refers to the period in days.

We can clearly see that the RL has outperformed the Equal Weight Portfolio during the testing period.

# FMCG SECTOR

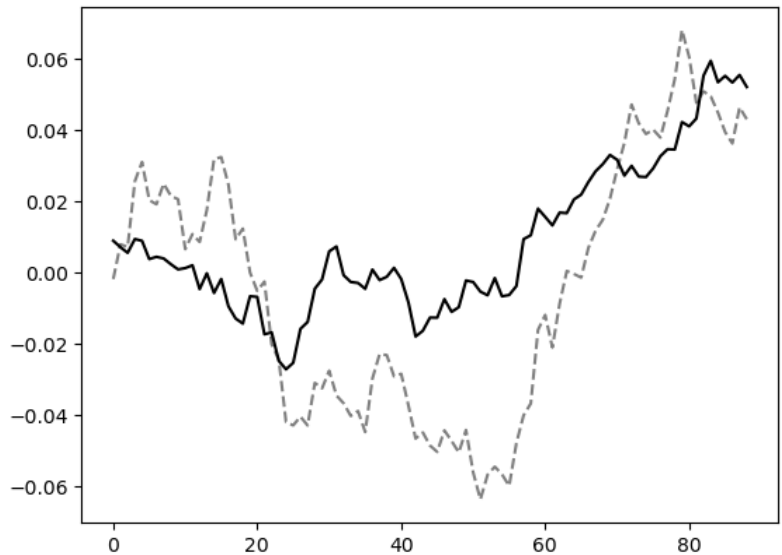

*Fig 27. Solid line refers to the performance of the RL portfolio and the dotted line refers to the performance of Equal weight portfolio for the testing period(03-Sep-2019 to 27-Feb-2020).*

In the y-axis the cumulative returns are plotted whereas the x-axis refers to the time period in days.

We can clearly see that the RL has outperformed the Equal Weight Portfolio for most of the period during testing.

## PHARMA SECTOR

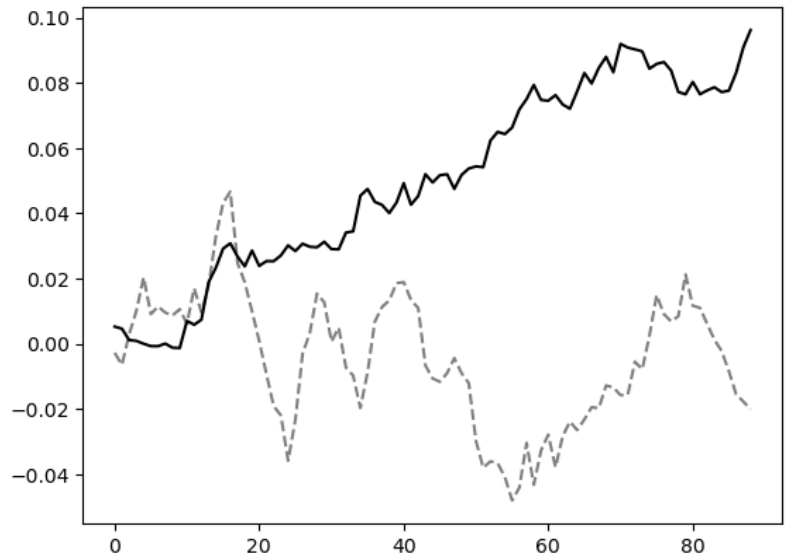

*Fig: 28: Solid line refers to the performance of the RL portfolio and the dotted line refers to the performance of the Equal weight portfolio for the testing period(03-Sep-2019 to 27-Feb-2020).*

In the y axis, the cumulative returns are plotted whereas the x axis refers to the period in days.

We can clearly see that the RL has outperformed the Equal Weight Portfolio during the testing period.

## BANKING SECTOR

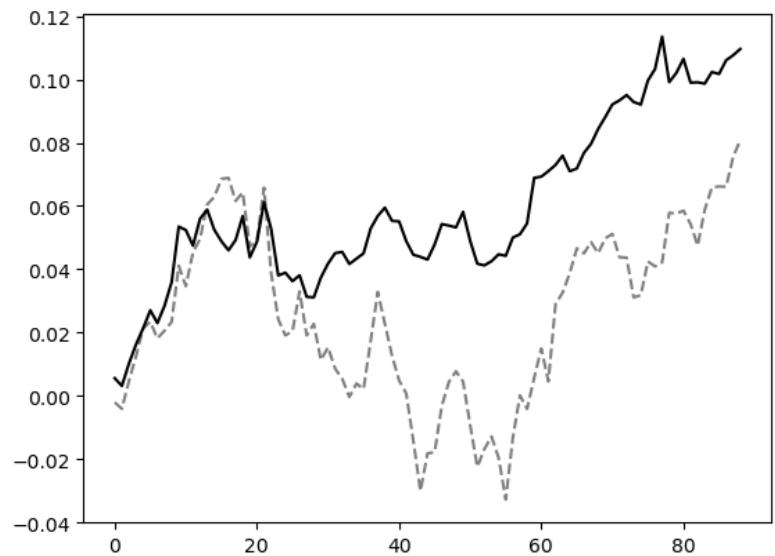

*Fig. 29. Solid line refers to the performance of the RL portfolio and the dotted line refers to the performance of the Equal weight portfolio for the testing period( 03-Sep-2019 to 27-Feb-2020).*

In the y-axis the cumulative returns are plotted whereas the x-axis refers to the time period in days.

We can clearly see that the RL has outperformed the Equal Weight Portfolio during the testing period.

## FINANCIAL SECTOR:

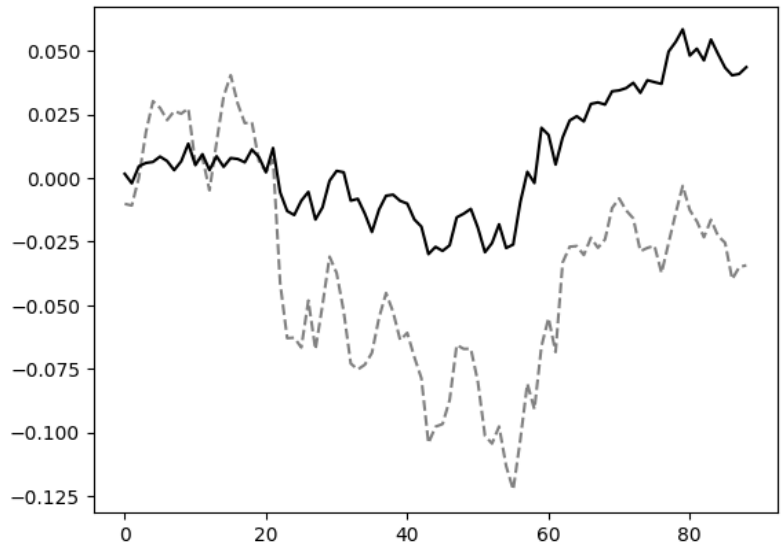

*Fig. 30. Solid line refers to the performance of RL portfolio and the dotted line refers to the performance of Equal weight portfolio for the testing period( 03-Sep-2019 to 27-Feb-2020).*

In the y axis the cumulative returns are plotted whereas the x-axis refers to the time period in days.

We can see that the RL has outperformed the Equal Weight Portfolio during the testing period

# IT SECTOR:

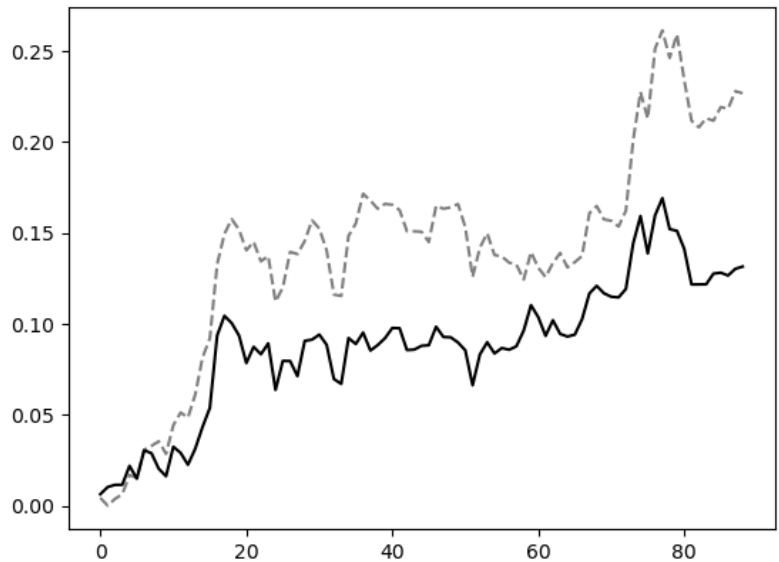

*Fig. 31. Solid line refers to the performance of the RL portfolio and the dotted line refers to the performance of the Equal weight portfolio for the testing period (03-Sep-2019 to 27-Feb-2020).*

In the y-axis the cumulative returns are plotted whereas the x-axis refers to the time period in days.

We can see that the Equal Weight Portfolio outperformed the RL Portfolio during the testing period.

## MIXED

For comparing the performance of RL and Equal Weight Portfolio we have chosen 2 stocks each from 5 different sectors which are as follows:

And the performance of the RL and Equal Weight Portfolio is as follows:

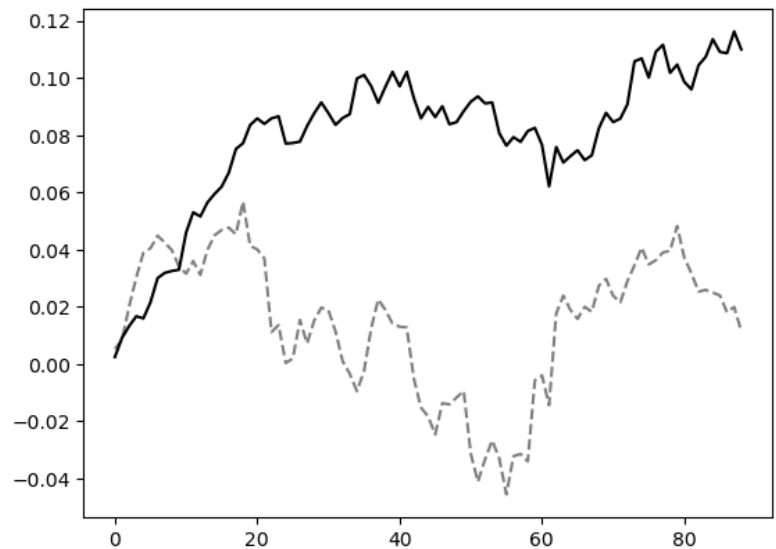

*Fig. 32. Solid line refers to the performance of the RL portfolio and the dotted line refers to the performance of the Equal weight portfolio for the testing period(03-Sep-2019 to 27-Feb-2020).*

In the y-axis the cumulative returns are plotted whereas the x-axis refers to the time period in days.

We can see that the RL Portfolio has outperformed the Equal Weight Portfolio during the testing period.

## Summary of Testing Results

Here is a summary of the performances of MVP, HRP, Equal Weight Portfolio, and RL Portfolio for the Out of sample data:

|  | MVP | HRP | EQUAL | RL |
|---|---|---|---|---|
| **Auto** | 1.245047 | 1.371209 | -0.6397 | 2.0812 |
| **Pharma** | 4.172059 | 4.071148 | -0.3732 | 4.3249 |
| **IT** | 1.483427 | 1.130776 | 3.0135 | 2.0895 |
| **Financial** | 0.646407 | 1.210418 | -0.3345 | 2.0895 |
| **Banking** | 1.9457 | 1.300964 | 1.577 | 3.3581 |
| **FMCG** | 1.633208 | 1.495954 | 1.0233 | 1.7219 |
| **Mixed** | 0.344013 | 0.943736 | 0.1248 | 3.4577 |

*Table 7: Comparison of the 4 different types of Portfolio Optimization Techniques.*

All the values mentioned in the table are the Sharpe ratios.

We can see that:

Apart from the IT sector in all the other sectors as well as for the Mixed portfolio the RL Portfolio Optimization Technique has yielded superior results w.r.t. the Sharpe ratio than all the other three Portfolio Optimization Techniques we have studied.

In the IT sector, RL has performed better than both HRP and MVP. But the Equal weight portfolio optimization technique has performed better than the RL Portfolio Optimization technique.

## 5. Conclusion

- From the Sharpe ratios, we can conclude that RL is capable of outperforming traditional portfolio optimization techniques like MVP & HRP.

- But much can't be said in the long run as the performance of RL is limited by computational capacity and volume of data.

- RL is dependent on the agent to learn; large datasets can be provided further to optimize portfolio based on the bullish or bearish performance of the stock at a time.

- The performance of the agent varies with different combinations of hyperparameters and gives optimal performance for a few selected combinations. These can be further tweaked for finding other maxima of performance.

## 6. Future Works & scope

- This project lays the groundwork for using reinforcement learning for the Indian stock market using Deep Q(N) Learning.

- Further building upon it, a more dynamic environment can be designed considering more factors, other indexes can be included, and other RL techniques can also be used.

[Type here]